\documentclass[10pt,twocolumn,letterpaper]{article}

 \usepackage{cvpr}              

\usepackage{cuted}
\usepackage{caption}
\usepackage{algorithm}
\usepackage{algpseudocode}
\usepackage{booktabs}

\definecolor{cvprblue}{rgb}{0.21,0.49,0.74}
\usepackage[pagebackref,breaklinks,colorlinks,allcolors=cvprblue]{hyperref}

\def\confName{CVPR}
\def\confYear{2027}

\title{Seeing Beyond Pixels: Radar Feature Generation from Images and Sparse Returns}

\author{%
Hongyu Zhou\textsuperscript{$^{1,2,3}$}\qquad
  Yubo Peng\textsuperscript{$^{1,2,3}$}\qquad
  Luping Xiang\textsuperscript{$^{1,2,3}$}
\thanks{Corresponding author:Luping Xiang(luping.xiang@nju.edu.cn)}\qquad
  Kun Yang\textsuperscript{$^{1,2,3}$}\\
  {\small \textsuperscript{$^1$}State Key Laboratory of Novel Software Technology, Nanjing University}
  {\small \textsuperscript{$^2$}Institute of Intelligent Networks and Communications (NINE)}\\
  {\small \textsuperscript{$^3$}School of Intelligent Software and Engineering, Nanjing University}\\[0.3em]
}

\begin{document}
\maketitle
\begin{strip}
  \centering
  \vspace*{-45pt}
  \raisebox{45pt}[\height][\depth]{
  \includegraphics[width=13cm]{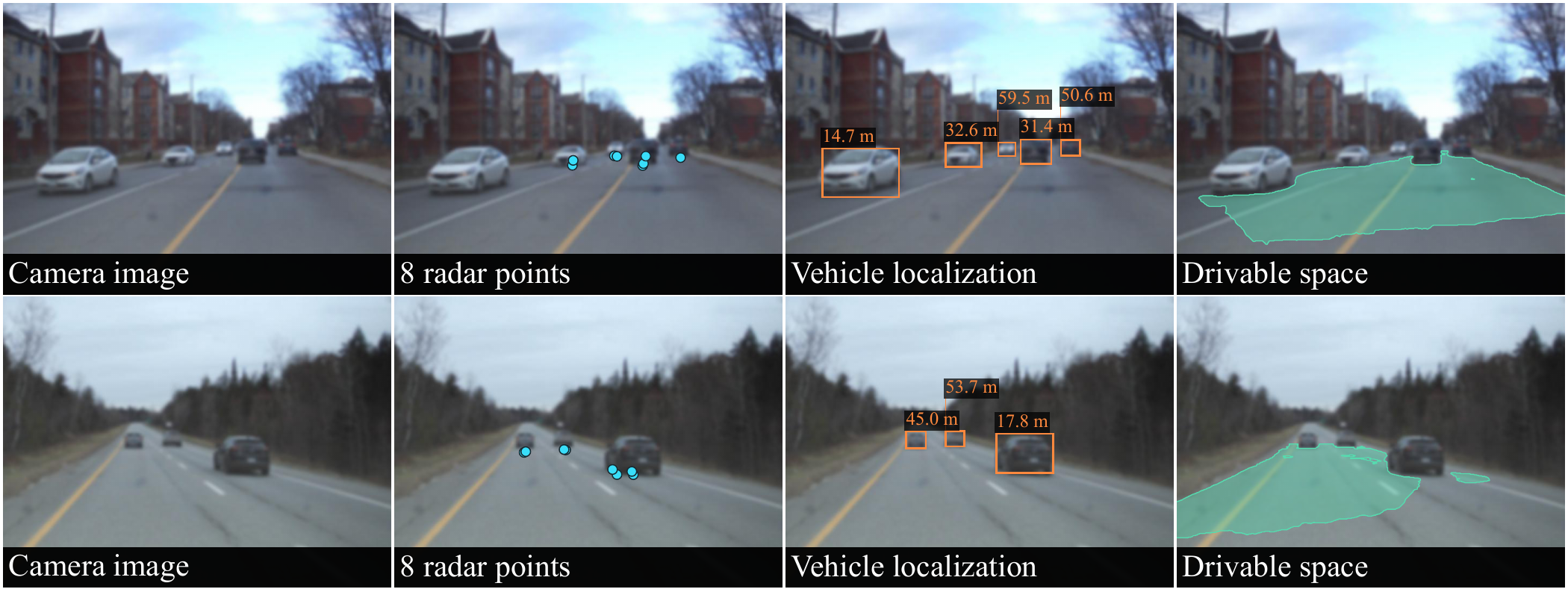}}
  \captionof{figure}{\textbf{VRTG} uses camera image and selected radar points to generate radar feature for fusion. Image-derived object cues guide how the radar measurements are used, while the fused features support vehicle localization with range estimates and drivable-space segmentation.}
  \label{fig:example}
\end{strip}
\label{sec:method}
\begin{abstract}
Camera-radar fusion benefits from complementary visual and geometric information but often relies on complete radar representations, leading to substantial computing overhead. However, the extent to which visual information can compensate for limited radar measurements remains insufficiently explored. Therefore, we propose Vision-informed Radar Token Generation (VRTG), which exploits cross-modal information to generate task-relevant radar features from camera images and sparse radar measurements. Specifically, VRTG first leverages image-derived object cues to select a limited number of informative radar points. It then establishes calibrated one-to-one associations between visual objects and radar returns, using matched measurements to refine object ranges while retaining visual estimates for unmatched objects. Finally, the associated object features and visual context are transformed into radar tokens compatible with existing camera-radar fusion networks. Experiments on RADIal demonstrate that VRTG achieves 90.76\% vehicle-detection F1 and 86.68\% free-space mIoU using only four radar points per frame, approaching the performance of the full-radar model. 
\end{abstract}  
\newpage
\section{Introduction}
\label{sec:intro}
Multimodal sensing integrates complementary information from different sensors to improve the accuracy and robustness of environmental perception \cite{11778081}. In particular, camera and millimeter-wave (mmWave) radar fusion is widely used in autonomous driving, intelligent transportation, and robotic perception \cite{10.1007/978-3-030-69535-4_24,singh2023depth,long2025riccardo,qiu2026rpgfusion}. Cameras provide rich visual information, including object appearance, texture, and semantic context, while mmWave radar offers geometric and motion-related measurements, such as range, azimuth, and radial velocity, with less sensitivity to illumination conditions. Their complementary capabilities support tasks such as object detection, tracking, depth estimation, and free-space segmentation. However, radar signals are typically acquired as raw analog-to-digital converter (ADC) samples and further processed into multidimensional spectra or point clouds. Retaining raw or intermediate radar representations increases the demands on data buffering, storage, signal processing, and feature extraction \cite{10.3389/fnins.2022.851774,paek2023enhanced,giroux2023t}.

To reduce radar processing overhead, recent studies have explored sample-wise and chirp-wise processing to reduce computational complexity and inference latency \cite{sen2026ssmradnet, Sen_2026_CVPR}. Meanwhile, radar-camera fusion systems often employ\newpage \noindent range-azimuth heatmaps or extracted radar point clouds to balance information retention and data size \cite{10.1007/978-3-030-69535-4_24,singh2023depth,long2025riccardo,qiu2026rpgfusion}. Other studies use image detections or previous radar observations to guide radar sample acquisition, suggesting that the value of radar measurements depends on the scene and perception task \cite{sakthi2021automotive,sakthi2022end,sakthi2023automotive}. Nevertheless, existing efficiency-oriented approaches mainly focus on reducing radar data through optimized acquisition, sampling, or processing, without fully exploiting the information shared between radar and vision. In particular, visual observations can provide object semantics and approximate spatial geometry, while radar measurements offer complementary physical information that is difficult to infer reliably from images alone. This raises an important question: Which radar information must be directly measured, and which can be inferred from visual observations? 

Motivated by this question, we propose VRTG (Vision-informed Radar Token Generation), which leverages image-derived object cues to associate sparse radar measurements with scene objects, thus generating radar feature tokens compatible with an existing camera-radar fusion network. These generated tokens replace the conventional radar feature input, enabling the fusion network to perform downstream tasks. Rather than reconstructing the original radar signals, VRTG focuses on generating task-relevant radar representations for multimodal perception (see \cref{fig:example}).

\noindent Our key contributions are as follows:
\begin{itemize}
    \item We propose VRTG, a vision-informed radar token generation framework that synthesizes radar features from images and sparse radar measurements. By exploiting cross-modal information, VRTG reduces the dependence on complete radar representations while preserving compatibility with existing camera-radar fusion networks.

    \item We develop vision-informed radar point selection and calibrated object-return association to prioritize task-relevant radar measurements and refine image-derived object geometry. This allows a limited number of radar points to provide complementary spatial information for downstream perception.

    \item We conduct extensive experiments on RADIal to evaluate the effects of radar point budgets and selection strategies. With only four radar points per frame, VRTG achieves 90.76\% detection F1 and 86.68\% free-space mIoU, approaching the full-radar model's performance and demonstrating the potential of cross-modal information to reduce radar input requirements.
\end{itemize}

\section{Related work}
\label{sec:related_work}

\subsection{Radar Representations for Multimodal Perception}
Radar perception commonly relies on raw signals, multidimensional spectra, or extracted point clouds, which offer different trade-offs between information retention and processing overhead. FFT-RadNet~\cite{Rebut_2022_CVPR} learns from range-Doppler (RD) spectra while avoiding the angle-estimation step used to construct denser radar views; EchoFusion~\cite{NEURIPS2023_a8f7f12b} uses range-time measurements in camera-radar fusion to retain information beyond detected points. Other work asks how much of a radar representation is needed for a task. SparseRadNet~\cite{10.1007/978-3-031-73016-0_4} retains 3.05\% of RD cells and only reduces detection F1 by 0.49 percentage points in its shared-backbone sampling experiment on RADIal~\cite{Rebut_2022_CVPR}. AdaRadar~\cite{park2026adaradar} reports a 101$\times$ reduction in transmitted RD data, with only 1.45-point F1 decrease under the same evaluation setting. These studies retain performance close to that of full-input models despite using limited radar data, suggesting that much of the complete radar representation is redundant for the evaluated tasks.

\subsection{Vision-Guided Radar Selection and Fusion}
Visual information provides object-level semantic and spatial cues that can guide the selection and interpretation of radar measurements. Previous studies have explored image-guided radar acquisition and detection-based subsampling to prioritize measurements relevant to scene objects. For example, CRN~\cite{10377206} uses radar points when lifting image features into bird's-eye view, while RaCFormer~\cite{11093090} and CVFusion~\cite{Zhong_2025_ICCV} develop query- and proposal-based mechanisms for relating the two modalities. R4Det~\cite{Xia_2026_CVPR} further combines radar-guided depth estimation with temporal and instance-level reasoning. For joint vehicle detection and free-space segmentation, REFNet++~\cite{11423030} instead fuses camera features with an RD-derived radar feature stream. TacoDepth~\cite{11092705} evaluates radar-camera depth estimation using 30 radar points. These systems establish several ways to use measured radar, but leave open how a fixed, small packet can supply the radar feature interface of an existing multi-task fusion network.

\subsection{Vision-Conditioned Radar Representation Generation}
Visual observations have also been used to predict radar representations. Alkanat and Pandharipande~\cite{10447724} estimate point-cloud density from images; RAIDS~\cite{10896467} and RAILS~\cite{11097772} estimate range-azimuth maps using visual cues together with depth or semantic descriptions. RadarGen~\cite{10.1007/978-3-032-37490-5_35} generates point clouds from multi-view images and evaluates their use in a downstream detector. When radar measurements are available, HGSFusion~\cite{Gu_Ma_Huang_Wei_Chen_Zhang_Hong_2025} uses image semantics to generate denser point sets for fusion. Together, these works show that visual cues can synthesize or densify radar representations for downstream perception. However, generating explicit point clouds or range-azimuth maps can increase model size and may reconstruct radar information that contributes little to downstream tasks.
\section{Method}
\label{sec:method}

\begin{figure*}[t]
  \centering
   \includegraphics[width=\textwidth]{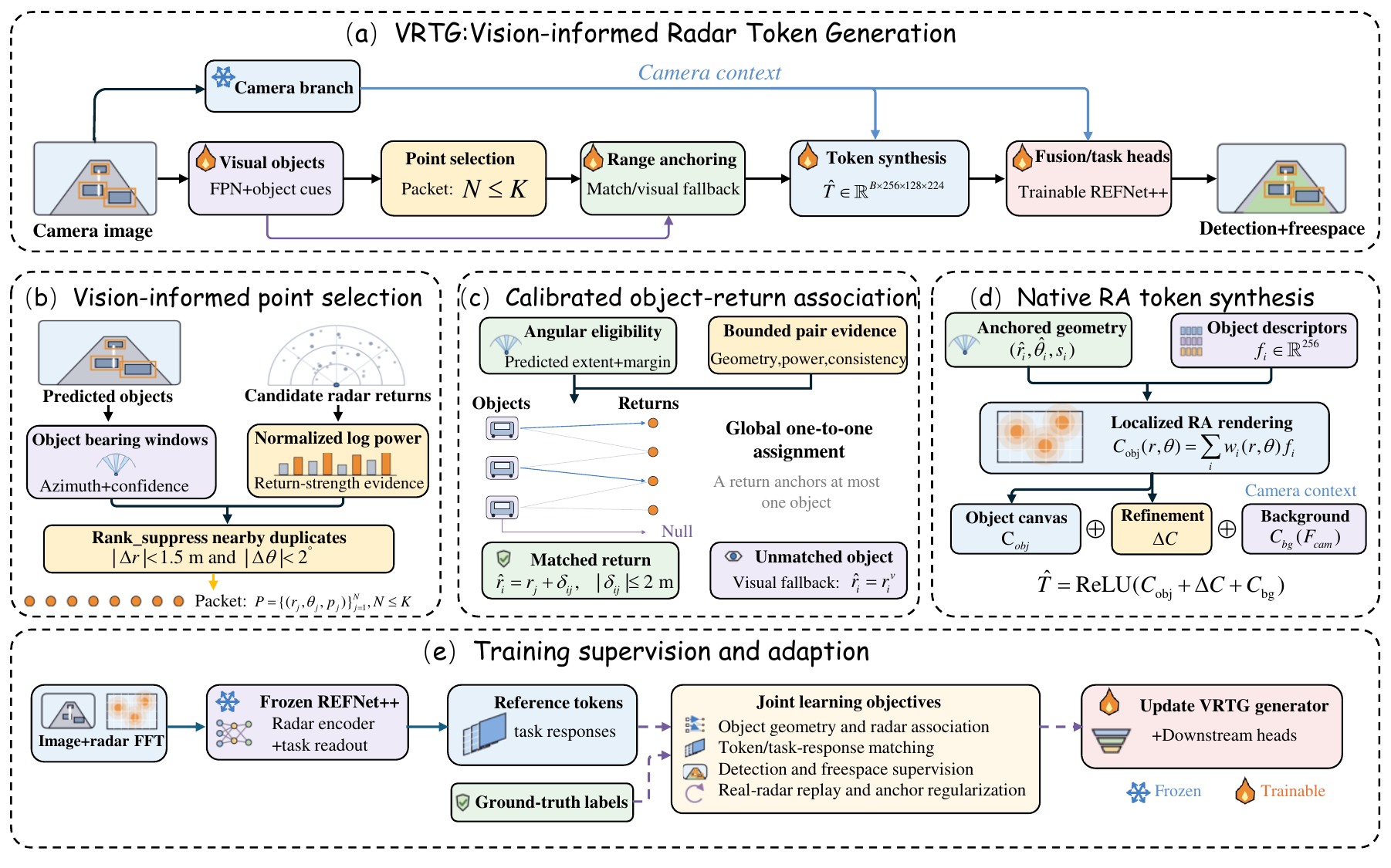}
   \caption{Method overview. (a) VRTG combines visual object cues with selected radar returns to synthesize radar tokens, which are fused with features from the camera branch for vehicle detection and free-space segmentation. (b) Visual object cues and return strength guide the selection of at most $K$ radar points. (c) One-to-one assignment anchors object ranges to matched returns, while unmatched objects retain visual estimates. (d) Localized rendering, refinement, and camera-conditioned background features form native radar tokens. (e) A frozen REFNet++~\cite{11423030} teacher and ground-truth labels supervise VRTG and the downstream heads.}
   \label{fig:architecture}
\end{figure*}

\Cref{fig:architecture} illustrates the overall architecture of VRTG, which comprises three main stages.
First, we extract object proposals and approximate spatial information from camera images to guide radar point selection. By considering visual relevance and radar return strength, we select at most \(K\) representative radar points while suppressing redundant measurements, as detailed in ~\cref{sec:3.1}.
Second, we establish one-to-one correspondences between visual objects and selected radar returns. Matched returns refine image-derived range estimates, while unmatched objects retain their visual estimates, allowing sparse radar measurements to complement visual geometry, as described in ~\cref{sec:3.2}.
Finally, we project the associated object features onto a range-azimuth grid and incorporate camera-derived contextual information to generate radar feature tokens. These tokens replace the conventional radar features in the downstream fusion network, enabling vehicle detection and free-space segmentation without requiring complete radar feature inputs, as presented in ~\cref{sec:3.3}.

\subsection{Vision-informed Radar Point Selection}\label{sec:3.1}
We select a compact radar packet using visual object cues and return strength~\cite{Gu_Ma_Huang_Wei_Chen_Zhang_Hong_2025}, with at most $K$ points retained per frame. An image feature pyramid predicts object boxes, confidence scores, and approximate ranges, from which we retain $M$ proposals at local confidence maxima~\cite{He_2016_CVPR}. Camera calibration maps these proposals to object bearings $\theta_i^v$ and angular half-widths $h_i$, providing spatial cues for ranking the radar candidates.

Let $\mathcal{P}=\{(r_j,\theta_j,p_j)\}_{j=1}^{L}$ denote the valid radar candidates, where each point contains range, azimuth, and power. For each candidate, we combine its angular relevance to the predicted objects with its normalized log power:
\begin{align}
    \ell_j &=\log\!\left(1+\max(p_j,0)\right), \\
    q_j &=\frac{\ell_j-\operatorname{median}(\boldsymbol{\ell})}{\max(\operatorname{std}(\boldsymbol{\ell}),\epsilon)}, \\
    a_j &=\max_i \left[s_i \exp\!\left(-\frac{1}{2}\left(\frac{\theta_j-\theta_i^v}{\max(h_i,h_{\min})}\right)^2\right)\right], \\
    \rho_j &= a_j+\lambda_p\,\operatorname{sigmoid}(q_j),
\label{eq:selection_score}
\end{align}
where $s_i$ is the confidence of proposal $i$, and $\rho_j$ is the final score of the candidate. The lower bounds $h_{\min}$ and $\epsilon$ stabilize angular scoring and power normalization, respectively, while $\lambda_p$ controls the contribution of return strength.

As summarized in ~\cref{alg:point_selection}, we repeatedly select the highest-scoring available return and suppress nearby candidates in both range and azimuth. This prevents a compact cluster from consuming the point budget, and preserves returns that are separated along either dimension. Selection stops when $K$ points have been chosen or no candidates remain, then the resulting packet $\mathcal{P}_K$ is passed to object-return association.

\begin{algorithm}[t]
\caption{Vision-informed radar point selection}
\label{alg:point_selection}
\begin{algorithmic}[1]
\Require Valid radar candidates $\mathcal{P}$; visual cues $\{(\theta_i^v,h_i,s_i)\}_{i=1}^{M}$; budget $K$; suppression thresholds $\tau_r,\tau_\theta$
\Ensure Selected packet $\mathcal{P}_K$, with $|\mathcal{P}_K|\leq K$
\State $\mathcal{P}_K \gets \varnothing$
\If{$K=0$ or $\mathcal{P}=\varnothing$}
    \State \Return $\mathcal{P}_K$
\EndIf
\State Compute $\rho_j$ for each candidate using
       Eq.~\eqref{eq:selection_score}
\State $\mathcal{A}\gets\{1,\ldots,L\}$
\While{$|\mathcal{P}_K|<K$ and $\mathcal{A}\neq\varnothing$}
    \State $j^\star\gets\arg\max_{j\in\mathcal{A}}\rho_j$
    \State $\mathcal{P}_K\gets \mathcal{P}_K\cup\{(r_{j^\star},\theta_{j^\star},p_{j^\star})\}$
    \State $\mathcal{D}\gets \{j\in\mathcal{A}:
           |r_j-r_{j^\star}|<\tau_r\;\land\;
           |\theta_j-\theta_{j^\star}|<\tau_\theta\}$
    \State $\mathcal{A}\gets\mathcal{A}\setminus\mathcal{D}$
\EndWhile
\State \Return $\mathcal{P}_K$
\end{algorithmic}
\end{algorithm}

\subsection{Calibrated Object-Return Association}
\label{sec:3.2}

Previous methods associate radar measurements with image-derived object centers~\cite{Nabati_2021_WACV} or image pixels~\cite{Long_2021_CVPR} to support geometric estimation. For a bounded radar packet, we formulate object-return association as a global one-to-one assignment with an unmatched option for each object.

For object $i$, let $r_i^v$, $\theta_i^v$, $h_i$, and $s_i$ denote its visual range, bearing, angular half-width, and confidence. Given a selected return $(r_j,\theta_j,p_j)$, we compute its normalized angular displacement $d_{ij}$:
\begin{equation}
d_{ij} = \frac{\theta_j-\theta_i^v}{\max(h_i,h_{\mathrm{a}})+m_\theta},
\end{equation}
where $h_{\mathrm{a}}$ provides a minimum angular width and $m_\theta$ adds a matching margin. A pair is eligible only when $|d_{ij}|\leq 1$ and the object confidence exceeds a threshold $\tau_s$.

We construct pair features $\mathbf{z}_{ij}$ from the normalized angular displacement, its magnitude, relative log power, object angular extent, and confidence. A small network $f_{\mathrm{a}}$ predicts a membership adjustment, which is then combined with angular alignment and visual-range consistency to calculate matching utility $u_{ij}$~\cite{Kim_Kim_Choi_Kum_2023}:
\begin{align}
u_{ij}&= 2-2d_{ij}^{2} + \beta\frac{f_{\mathrm{a}}(\mathbf{z}_{ij})} {1+|f_{\mathrm{a}}(\mathbf{z}_{ij})|} - c_{ij}, \\
c_{ij}&= \gamma\left[1-\exp\left(-\frac{(r_j-r_i^v)^2}{2\sigma_i^2}\right)\right],
\qquad
\sigma_i=\sigma_0+\sigma_1 r_i^v,
\end{align}
where $\beta$ bounds the learned adjustment, while $\gamma$ limits the range-consistency penalty. The latter saturates as range disagreement increases, preventing visual-range errors from imposing an unbounded penalty on otherwise plausible matches. Absolute ranges and range differences enter through $c_{ij}$ rather than the learned pair features.

Let $X_{ij}\in\{0,1\}$ indicate whether radar return $j$ is assigned to object $i$, each object has a zero-score unmatched option. We optimize the assignment matrix $\mathbf{X}$ to maximize the total matching utility, subject to at most one return per object and one object per return:
\begin{equation}
\begin{aligned}
\max_{\mathbf{X}\in\{0,1\}^{M\times N}}&\quad \sum_{i=1}^{M}\sum_{j=1}^{N}X_{ij}u_{ij},\\
\text{subject to}
&\quad \sum_j X_{ij}\leq 1,\qquad \sum_i X_{ij}\leq 1 .
\end{aligned}
\end{equation}
We solve the resulting problem using Hungarian assignment with an individual null column for each object~\cite{hungarian}. Consequently, a return can anchor at most one object, and objects without a beneficial assignment remain unmatched.

For a matched pair, we define $\delta_{ij}$ as a signed distance correction, a gated correction adjusts the measured range within a prescribed bound~\cite{Long_Kumar_Morris_Liu_Castro_Chakravarty_2023}:
\begin{align}
\delta_{ij} &= \delta_{\max}\tanh\!\left(f_{\delta}(\mathbf{z}_{ij})\right)\operatorname{sigmoid}\!\left(f_{\mathrm{g}}(\mathbf{z}_{ij})\right),\\
\hat r_i&=\begin{cases}r_j+\delta_{ij}, & X_{ij}=1,\\
r_i^v, & \sum_j X_{ij}=0 .
\end{cases}
\end{align}
Thus, matched objects use a measured range with a bounded adjustment, while unmatched objects preserve their visual estimates. The resulting geometry guides the spatial placement of object features during radar token generation.

\begin{table*}[t]
  \centering
  \caption{Vehicle-detection results reported on RADIal. RD, ADC, RT, PC, and C denote range-Doppler, analog-to-digital converter signals, range-time, point cloud, and front-view camera data, respectively. Single-task and multi-task refer to detection-only and joint detection-free-space-segmentation settings. AP and AR under the original RADIal protocol average precision and recall over confidence thresholds. PointTransformerV3's RADIal values are reported by ~\cite{Alsharif_2026_CVPR}. - indicates an unreported or unavailable result. $^*$ Trained with visual-guided strategy. }
  \label{table:radial-recent-results}
  \scriptsize
  \setlength{\tabcolsep}{3pt}
  \renewcommand{\arraystretch}{1.15}
  \begin{tabular*}{\textwidth}{@{\extracolsep{\fill}}ll*{10}{c}@{}}
    \toprule
    & & \multicolumn{5}{c}{Single-task} & \multicolumn{5}{c}{Multi-task} \\
    \cmidrule(lr){3-7}\cmidrule(l){8-12}
    Methods & Input
      & AP(\%)$\uparrow$ & AR(\%)$\uparrow$ & F1(\%)$\uparrow$ & RE(m)$\downarrow$ & AE($^\circ$)$\downarrow$
      & AP(\%)$\uparrow$ & AR(\%)$\uparrow$ & F1(\%)$\uparrow$ & RE(m)$\downarrow$ & AE($^\circ$)$\downarrow$ \\
    \midrule
    FFT-RadNet~\cite{Rebut_2022_CVPR} & RD
      & 93.45 & 83.35 & 88.11 & 0.12 & 0.15 & 96.05 & 82.18 & 88.57 & \textbf{0.11} & 0.17 \\
    EchoFusion~\cite{NEURIPS2023_a8f7f12b} & RT+C
      & \textbf{96.95} & \textbf{93.43} & \textbf{95.16} & 0.12 & 0.18 & - & - & - & - & - \\
    EFA-RadNet~\cite{s26072050} & RD
      & - & - & - & - & - & \textbf{97.22} & \textbf{90.42} & \textbf{93.70} & 0.12 & 0.10 \\
    T-FFTRadNet~\cite{giroux2023t} & ADC
      & 90.80 & 88.31 & 89.54 & 0.15 & 0.13 & 89.60 & 89.50 & 89.50 & 0.15 & 0.13 \\
    SSMRadNet~\cite{sen2026ssmradnet} & ADC
      & - & - & - & - & - & 83.00 & 71.00 & 77.00 & 0.14 & 0.15 \\
    SpectralPillars~\cite{Alsharif_2026_CVPR} & spectral PC
      & 94.30 & 91.50 & 92.90 & 0.12 & 0.22 & - & - & - & - & - \\
    PointTransformerV3~\cite{Wu_2024_CVPR} & spectral PC
      & 96.60 & 92.70 & 94.60 & \textbf{0.10} & 0.12 & - & - & - & - & - \\
    ADCNet~\cite{yang2023adcnetlearningrawradar} & ADC
      & - & - & - & - & - & 95.00 & 89.00 & 91.90 & 0.13 & 0.11 \\
    REFNet++~\cite{11423030} & RD+C
      & 96.92 & 90.68 & 93.70 & 0.11 & \textbf{0.09} & 96.16 & 89.43 & 92.67 & \textbf{0.11} & \textbf{0.09} \\
    \midrule
    VRTG$^*$ & C + $4$ radar points
      & 90.92 & 85.97 & 88.38 & 0.225 & 0.209 & 92.85 & 88.76 & 90.76 & 0.222 & 0.211 \\
    VRTG$^*$ & C + $8$ radar points
      & 92.11 & 87.69 & 89.85 & 0.224 & 0.212 & 91.13 & 87.91 & 89.49 & 0.220 & 0.211 \\
    \bottomrule
  \end{tabular*}
\end{table*}

\subsection{Native Radar Token Generation}
\label{sec:3.3}

We convert the associated object geometry and visual features into a spatial representation that matches the radar-token interface of the downstream model. For matched objects, the updated range is used to recompute the calibrated bearing while retaining the visual angular residual. For unmatched objects, simply retaining the image-predicted geometry is enough. The positional component of each object descriptor is updated accordingly, so the feature encoding follows its spatial placement.

Let $\mathbf{f}_i\in\mathbb{R}^{256}$ denote the resulting descriptor of object $i$, with range $\hat r_i$, bearing $\hat\theta_i$, and confidence $s_i$. We place these descriptors on a range-azimuth grid $C_{\mathrm{obj}}$ using Gaussian weights $w_i(r,\theta)$~\cite{11094837,10.1007/978-3-032-37010-5_18}:
\begin{align}
w_i(r,\theta)&= s_i\exp\left[-\frac{1}{2}\left(\frac{(r-\hat r_i)^2}{\Delta r^2}+\frac{(\theta-\hat\theta_i)^2}{\Delta\theta^2}\right)\right],\\
C_{\mathrm{obj}}(r,\theta)&= \sum_{i=1}^{M}w_i(r,\theta)\mathbf{f}_i,
\end{align}
where $\Delta r$ and $\Delta\theta$ are the grid resolutions. The Gaussian bandwidth is fixed at one cell along each axis, allowing continuous object coordinates to control feature placement. Both matched and unmatched proposals contribute to the canvas according to their confidence.

A convolutional refiner then predicts a residual feature map $U$ from the object canvas and its confidence map $S(r,\theta)=\max_i w_i(r,\theta)$. We constrain the residual magnitude relative to the local object features:
\begin{equation}
\Delta C(r,\theta)=U(r,\theta)\min\left(1,\frac{\eta L(r,\theta)}{\max(\operatorname{RMS}_{c}(U(r,\theta)),\epsilon)}\right),
\end{equation}
where $\operatorname{RMS}_{c}$ computes the root mean square over channels. The local scale $L$ is obtained by max pooling the channel RMS of $C_{\mathrm{obj}}$ over an $11\times11$ neighborhood, with a small positive floor. This bounds the refinement $\Delta C(r,\theta)$ while allowing it to adjust the rendered features.

To provide context beyond the object-centered regions, a separate background path projects a coarse version of the frozen camera features $F_{\mathrm{cam}}$ and upsamples it to the range-azimuth grid. The final radar tokens $\hat T$ are:
\begin{equation}
\hat T = \operatorname{ReLU}\left(C_{\mathrm{obj}}+\Delta C+C_{\mathrm{bg}}(F_{\mathrm{cam}})\right).
\end{equation}
The generated radar tokens follow downstream model's native radar feature interface, allowing them to enter its fusion stage in place of features extracted from the full radar input. The fusion stage combines these tokens with camera features, which are used by downstream attention heads. In this way, the model retains the original perception tasks while using a limited set of measured radar returns to construct its radar features.

\section{Experiments}

\label{sec:experiments}
\begin{table}[t]
  \centering
  \caption{Free-space segmentation results reported on RADIal.}
  \label{table:radial-freespace}
  \scriptsize
  \begin{tabular*}{\linewidth}{@{\extracolsep{\fill}}llc@{}}
    \toprule
    Method & Input & mIoU (\%) $\uparrow$ \\
    \midrule
    FFT-RadNet~\cite{Rebut_2022_CVPR} & RD & 74.00 \\
    T-FFTRadNet (ADC)~\cite{giroux2023t} & ADC & 79.60 \\
    T-FFTRadNet (RD)~\cite{giroux2023t} & RD & 80.20 \\
    ADCNet~\cite{yang2023adcnetlearningrawradar} & ADC & 78.59 \\
    SparseRadNet~\cite{10.1007/978-3-031-73016-0_4} & RD & 78.48 \\
    SSMRadNet~\cite{sen2026ssmradnet} & ADC & 79.00 \\
    RAVEN~\cite{Sen_2026_CVPR} & ADC & \textbf{90.00} \\
    SSRaDNet (RD)~\cite{s26196074} & RD & 80.60 \\
    SSRaDNet (ADC)~\cite{s26196074} & ADC & 79.89 \\
    REFNet++~\cite{11423030} & RD+C & 87.58 \\
    \midrule
     VRTG$^*$ & C+4 radar points & 86.68 \\
     VRTG$^*$ & C+8 radar points & 86.67 \\
    \bottomrule
  \end{tabular*}
\end{table}

\begin{figure*}[t]
  \centering
   \includegraphics[width=\textwidth]{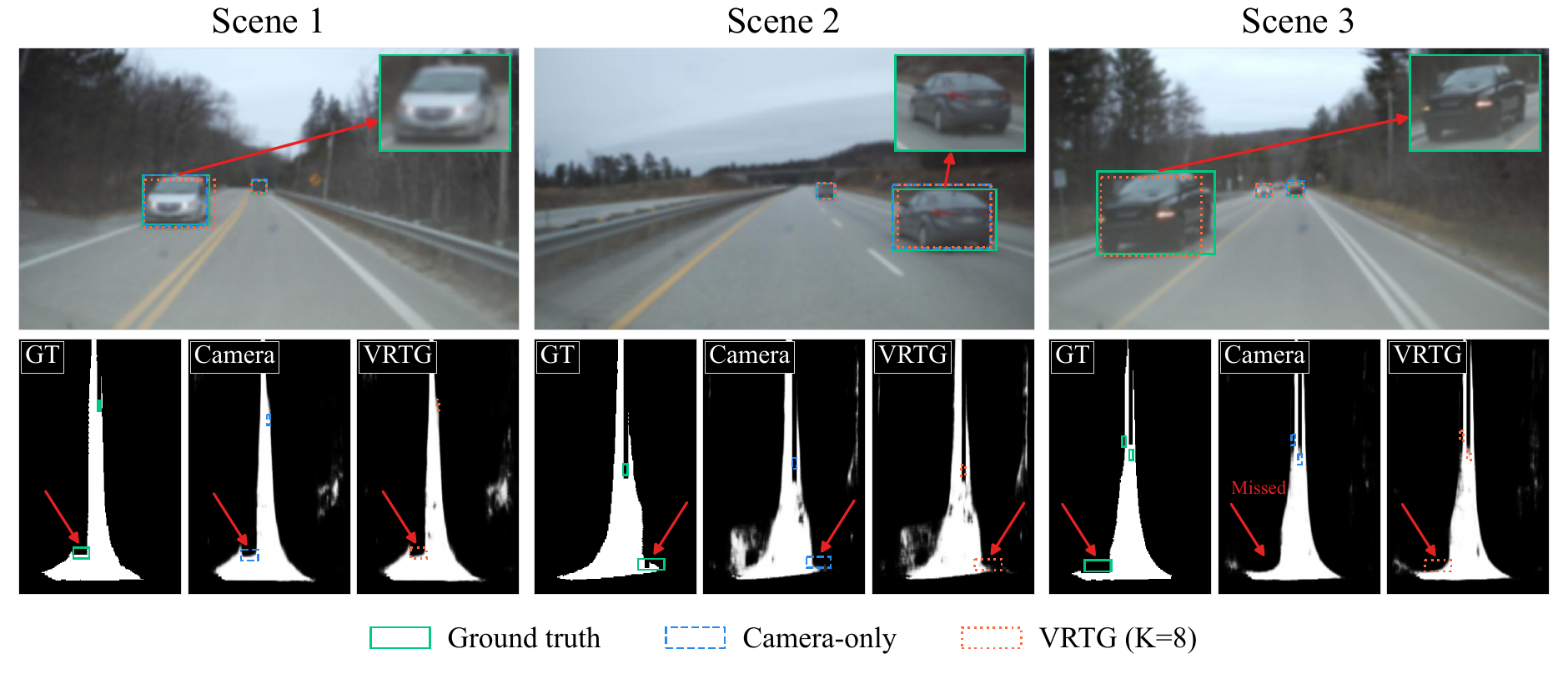}
   \caption{Qualitative results of VRTG on RADIal. Each column shows one scene: the front-view image appears above the ground-truth, camera-only, and VRTG polar-view maps, arranged from left to right. VRTG uses $K=8$ visually selected radar points per frame. Insets enlarge selected vehicles, and arrows highlight representative detections. In the polar-view maps, white denotes free space and black denotes background.}
   \label{fig:visual}
\end{figure*}

To evaluate the effectiveness of VRTG in multimodal perception with limited radar measurements, we conduct experiments on the RADIal dataset for vehicle detection and free-space segmentation. We first compare VRTG with existing methods and the full-radar REFNet++ model. We then present qualitative results (\cref{subsec:1}), investigate the effects of radar point budgets and selection strategies (\cref{subsec:2}), and analyze detection performance across target ranges, azimuth angles, and scene densities (\cref{subsec:3}). All experiments are conducted on a server equipped with four NVIDIA GeForce RTX 5090 GPUs.

\vspace{4pt}
\noindent\textbf{Dataset and teacher model.}
We evaluate VRTG on RADIal~\cite{Rebut_2022_CVPR}, an automotive perception dataset containing synchronized high-definition radar, front-view camera, and LiDAR recordings from urban, rural, and highway driving. It comprises approximately 25,000 synchronized frames, of which 8,252 are annotated with 9,550 vehicles. Vehicle locations are provided in radar range and azimuth, together with free-space segmentation masks. We use the predefined sequence-wise training, validation, and test split. The pretrained REFNet++~\cite{11423030}, which takes camera images and range-Doppler spectra as input and jointly predicts vehicle detections and free space, serves as the teacher model. 

\vspace{4pt}
\noindent\textbf{Evaluation metrics.}
We evaluate vehicle detection using average precision (AP), average recall (AR), and their harmonic mean $F_1$. Predictions are matched to ground-truth vehicles at an intersection-over-union threshold of $0.5$. Following the RADIal evaluation protocol, AP and AR are the mean precision and recall over nine confidence thresholds from $0.1$ to $0.9$, rather than precision-recall curve area. We additionally report range error (RE, in meters) and azimuth-angle error (AE, in degrees) for matched detections. Free-space segmentation is evaluated by mean intersection over union (mIoU) within $0$-$50\,\mathrm{m}$, following the REFNet++ protocol.
\begin{figure*}[t]
  \centering
   \includegraphics[width=\textwidth]{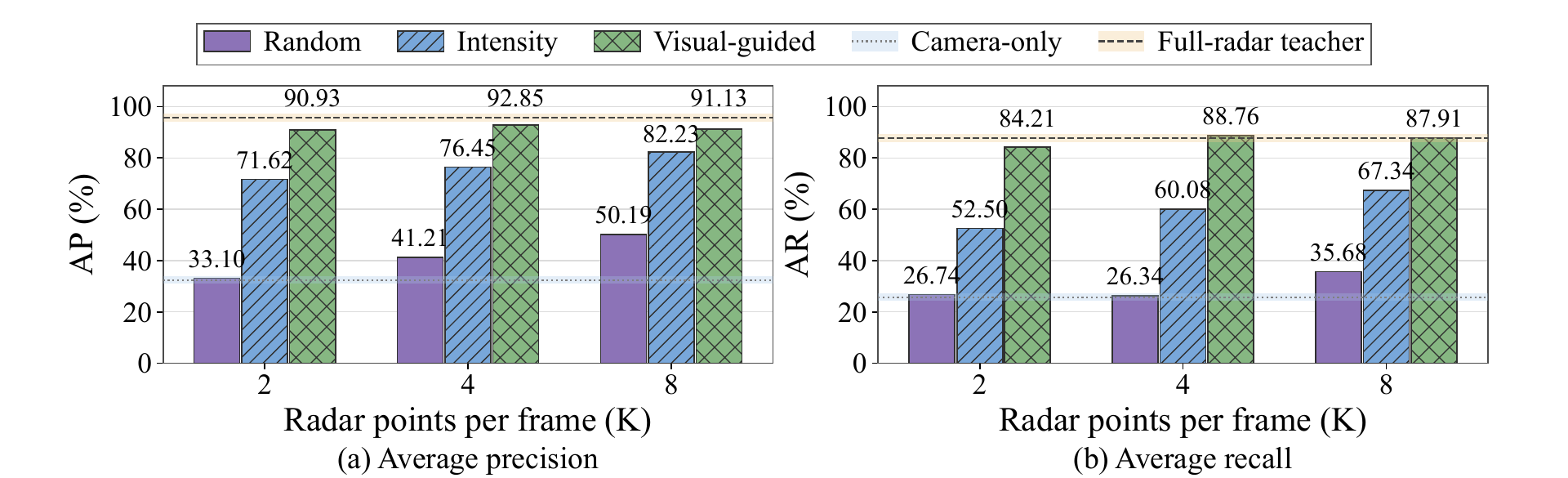}
   \caption{Detection AP and AR with $K=2, 4,$ and $8$ radar points per frame under random, intensity-based, and visual-guided selection, each trained individually. Dotted and dashed lines indicate the camera-only and full-radar teacher results, respectively.}
   \label{fig:01}
\end{figure*}
\begin{figure*}[t]
  \centering
   \includegraphics[width=\textwidth]{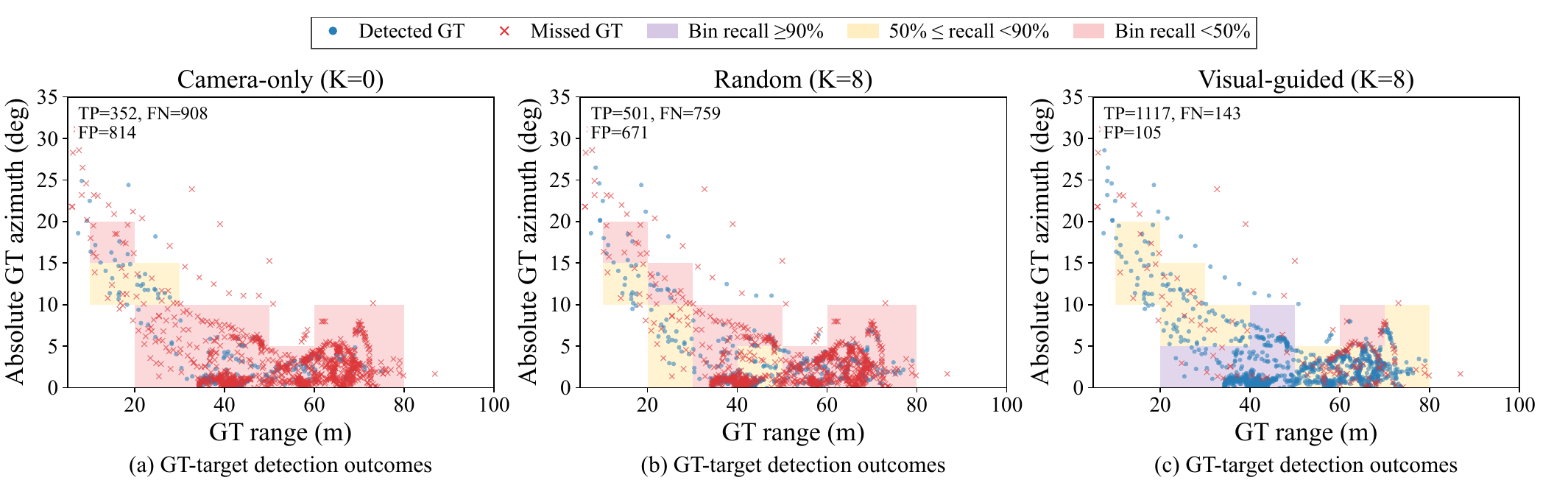}
   \caption{Detected and missed ground-truth vehicles across range and azimuth for camera-only, random, and visual-guided selection.}
   \label{fig:03}
\end{figure*}
\begin{figure*}[t]
  \centering
   \includegraphics[width=\textwidth]{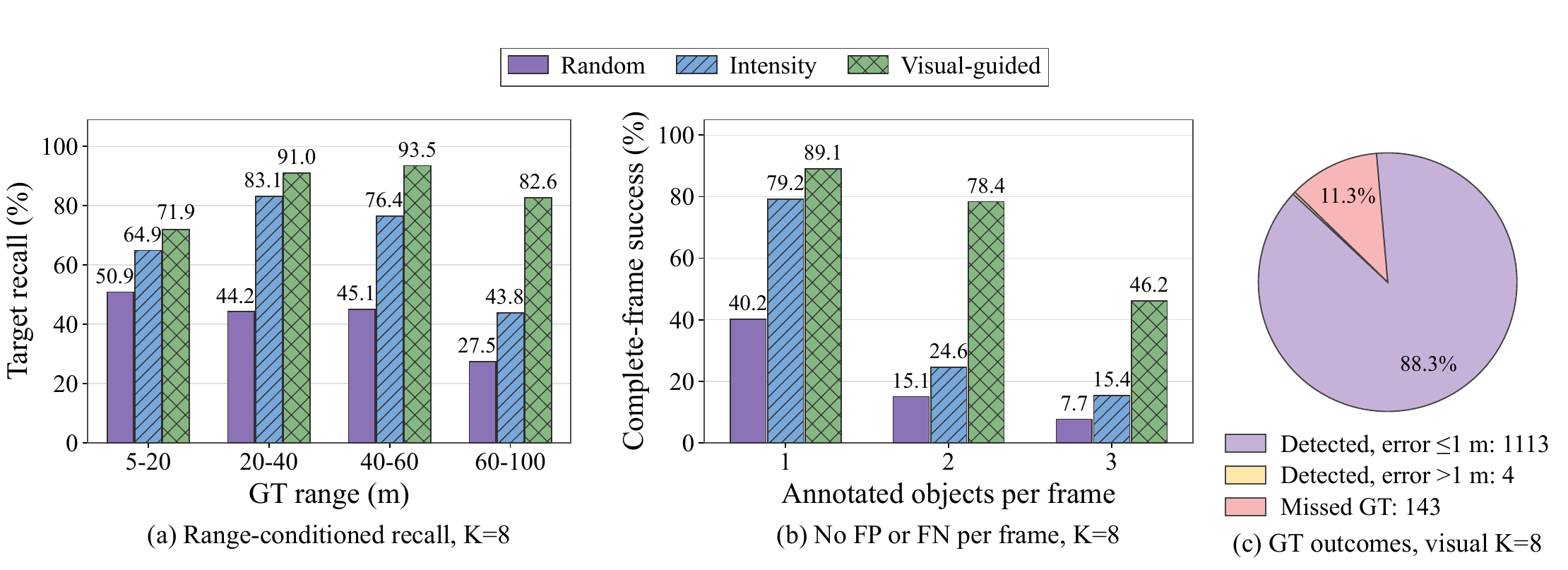}
   \caption{Detection results across target range and scene density, together with the outcome breakdown for visual-guided selection.}
   \label{fig:05}
\end{figure*}

\subsection{Qualitative Analysis}
\label{subsec:1}
~\cref{table:radial-recent-results} and ~\cref{table:radial-freespace} show that the gap between VRTG and the teacher model REFNet++~\cite{11423030} is more pronounced in vehicle localization than in free-space segmentation. In the multi-task setting, VRTG achieves F1 scores of 90.76\% and 89.49\% with four and eight selected radar points, respectively, compared with 92.67\% for REFNet++. The range and azimuth errors are about $0.22\,\mathrm{m}$ and $0.21^\circ$ for both point budgets, whereas REFNet++ reports $0.11\,\mathrm{m}$ and $0.09^\circ$. One possible explanation is that VRTG uses matched radar returns to anchor object ranges, while objects without a match retain their visual range estimates. Because bearings are derived primarily from image geometry, even objects with a radar range anchor still depend on visual cues for azimuth. This reliance on visual geometry may partly explain the higher range and azimuth errors relative to REFNet++, which derives range-azimuth features from dense range-Doppler measurements. For free-space segmentation, VRTG achieves 86.68\% and 86.67\% mIoU with four and eight points, respectively, within 0.91 percentage points of REFNet++. The near-identical scores across the two budgets are consistent with the camera branch providing scene-level cues about road layout and boundaries, even when radar measurements are sparse.

\Cref{fig:visual} illustrates the difference between camera-only and VRTG predictions. The camera-only model can identify visible vehicles, but its detections do not always align with their annotated positions in the polar view. With $K=8$ visually selected radar points, VRTG places vehicle detections closer to the annotated range and azimuth in these examples. The nearby vehicle on the right in Scene 1 makes this difference particularly clear. The visual comparison suggests that sparse radar measurements can help constrain vehicle location and range after the image provides appearance cues.

\subsection{Ablation Studies}
\label{subsec:2}

We ablate the radar-point budget and selection strategy in ~\cref{fig:01}. The camera-only setting ($K=0$) and the full-radar teacher provide references for interpreting the results.

\vspace{4pt}
\noindent\textbf{Selection strategy.} At the same point budget, visual-guided selection consistently outperforms random and intensity-based selection. With $K=2$, it achieves 90.93\% AP and 84.21\% AR, compared with 71.62\% and 52.50\% for intensity-based selection, and 33.10\% and 26.74\% for random selection. A plausible explanation is that random sampling ignores object locations, whereas intensity ranking may spend several slots on strong returns from one reflector. Therefore, our strategy combines return strength with confidence-weighted agreement to camera-predicted object bearings and suppresses nearby duplicates in range and angle, which encourages the limited packet to cover more image-supported regions. Visual-guided selection peaks at $K=4$, reaching 92.85\% AP and 88.76\% AR, and the advantage persists at $K=8$, exceeding intensity-based selection by 8.90 AP points and 20.57 AR points. However, at $K=8$, AP falls to 91.13\% and AR to 87.91\%, decreases of 1.72 and 0.85 percentage points, respectively. This suggests that the four additional slots may admit weaker returns or points with ambiguous object associations after the most relevant measurements have been selected, potentially perturbing range estimation without improving object coverage.

\vspace{4pt}
\noindent\textbf{Effect of selected radar points.} The camera-only reference shows performance without radar measurements, which only achieves 32.95\% AP and 26.29\% AR.  Even adding two randomly selected points  only yields 33.10\% AP and 26.74\% AR, remaining close to that reference. Thus, providing a small radar packet alone does not ensure a useful detection gain. With the same two-point budget, visual-guided selection reaches 90.93\% AP and 84.21\% AR, indicating that the measurements contribute most when the packet prioritizes returns associated with visible object candidates. 

\subsection{Detection across Range and Azimuth}
\label{subsec:3}
\Cref{fig:03} shows where detection outcomes differ across the range-azimuth plane. With $K=8$, visual-guided selection detects 1,117 of 1,260 annotated vehicles, with 143 misses and 105 false positives. Random selection detects 501 vehicles, with 759 misses and 671 false positives, while the camera-only setting detects 352, with 908 misses and 814 false positives. The detected targets under visual-guided selection extend across a broader range of distances. The underlying mechanism of this difference is that random sampling may spend the limited point budget on returns unrelated to vehicles, whereas visual guidance prioritizes returns near predicted object bearings.

The range-conditioned results in \cref{fig:05}(a) show that the gain of the sampling is not uniform over range. Visual-guided recall reaches 91.0\% at 20-40\,m and 93.5\% at 40-60\,m. At 60-100\,m, it remains at 82.6\%, compared with 43.8\% for intensity-based and 27.5\% for random selection, which indicates that selecting the strongest returns alone therefore does not reliably preserve the measurements most useful for distant vehicles. Visual-guided recall decreases to 71.9\% below 20\,m and 82.6\% beyond 60\,m. This decrease may be attributed to different range-dependent factors, nearby vehicles can span a wide angular region or be partially outside the image, complicating radar-image association, whereas distant vehicles occupy fewer pixels and may produce weaker radar returns.

~\Cref{fig:05}(b) further shows that complete-frame success declines as the number of annotated vehicles increases. For visual-guided selection, it falls from approximately 89\% in single-object frames to 46\% in three-object frames, but remains above the other strategies. With a fixed per-frame budget, more objects must share the selected returns, increasing the difficulty of associating a useful measurement with each vehicle. The all-or-nothing frame metric also creates more opportunities for a single miss or false positive to make a frame unsuccessful. \cref{fig:05}(c) separates the 1,260 annotated vehicles into three outcomes under visual-guided selection at $K=8$. Among them, 1,113 (88.3\%) are detected with an absolute range error of at most 1\,m, only four (0.3\%) are detected with a larger error, and 143 (11.3\%) are missed. Thus, 99.6\% of the detected vehicles meet the 1\,m range criterion. Most remaining failures are therefore missed detections rather than large range errors for detected vehicles. This suggests that covering and associating all vehicles with a fixed per-frame point budget is more challenging than estimating the range of vehicles that have been detected, which remains a promising direction for future work.

\section{Conclusion}
\label{sec:conclusion}
We presented VRTG, a method that generates radar tokens from a camera image and a fixed budget of selected radar points. By using visual predictions to prioritize radar returns and measured returns to constrain object range, VRTG supplies the REFNet++'s detection and free-space heads with features in their native radar representation. Our analyses on RADIal show that visual-guided selection yields substantially higher detection AP and AR than random or intensity-based selection at the same point budget. 
\newpage
\section{LLM Usage Statement}
ChatGPT (OpenAI) was used to assist with implementing and debugging experimental code, summarizing related work, and drafting and editing English manuscript text. But the research idea, method, and experimental design were all determined by the authors without LLM participation. All experimental measurements and numerical results were produced by the authors' experiments, and the plots of these results do not contain LLM-generated data. The authors reviewed and verified all AI-assisted content and take full responsibility for the final work.
\newpage
{
    \small
    \bibliographystyle{ieeenat_fullname}
    \bibliography{main}
}


\end{document}